\documentclass{article} 
\usepackage{iclr2019_conference,times}

\usepackage{amsmath,amsfonts,bm}

\def\eqref#1{equation~\ref{#1}}

\def\1{\bm{1}}

\DeclareMathAlphabet{\mathsfit}{\encodingdefault}{\sfdefault}{m}{sl}
\SetMathAlphabet{\mathsfit}{bold}{\encodingdefault}{\sfdefault}{bx}{n}

\usepackage{hyperref}
\usepackage{url}
\usepackage{xargs}                      
\usepackage{ifdraft, blindtext}
\usepackage{todonotes}
\usepackage{csquotes}
\usepackage{comment}

\title{The Scientific Method in the Science of Machine Learning}

\author{Jessica Zosa Forde\\
Project Jupyter\\
\texttt{jzf2101@columbia.edu}\\
\And
Michela Paganini \\
Facebook AI Research \\
\texttt{michela@fb.com} \\
}

\iclrfinalcopy 
\begin{document}

\maketitle

\begin{abstract}
In the quest to align deep learning with the sciences to address calls for rigor, safety, and interpretability in machine learning systems, this contribution identifies key missing pieces: the stages of hypothesis formulation and testing, as well as statistical and systematic uncertainty estimation -- core tenets of the scientific method. 
This position paper discusses the ways in which contemporary science is conducted in other domains and identifies potentially useful practices. We present a case study from physics and describe how this field has promoted rigor through specific methodological practices, and provide recommendations on how machine learning researchers can adopt these practices into the research ecosystem.
We argue that both domain-driven experiments and application-agnostic questions of the inner workings of fundamental building blocks of machine learning models ought to be examined with the tools of the scientific method, to ensure we not only understand effect, but also begin to understand cause, which is the raison d'\^{e}tre of science.
\end{abstract}

\section{Introduction}
As machine learning (ML) matures and ensconces itself in the daily lives of people, calls for assurances of safety, rigor, robustness, and interpretability have come to the forefront of discussion. Many of these concerns, both social and technical, spring from the lack of understanding of the root causes of performance in ML systems~\citep{Rahimi2017-gt, Lipton2018-lm, Sculley2018-le, Sculley2015-da, Woods2018-di}.
Researchers have emphasized the need for more thorough examination of published results. Replications of previous work in GANs \citep{Lucic2018-bp}, language modeling \citep{Melis2018-mv, Vaswani2017-iu}, Bayesian neural networks \citep{Riquelme2018-cn}, reinforcement learning \citep{Henderson2018-yx, Mania2018-dt, Nagarajan2019-vd} that attempted to understand if extensions to simpler models affect performance have yielded negative results. Large scale studies of published work have attempted to provide the community with independent re-implementations of machine learning experiments \citep{Pineau2017-ya, Tian2019-uk}. At the same time, researchers have demonstrated that trained models fail to generalize to novel scenarios \citep{Recht2019-km, Zhang2019-cd, Zhang2018-bq} and suggested that this decrease in performance may be due to confounders in training data \citep{Stock2018-py, Zhang2017-so}.  These inconsistencies in performance are particularly troubling in deployed systems, whose errors have real world implications \citep{Buolamwini2018-lt, Rudin2018-rw, Zech2018-iw}.  Many have identified various social factors within the research ecosystem that contribute to these challenges \citep{Lipton2018-lm, Sculley2018-or, Sculley2018-le, Henderson2018-km, Gebru2018-rh, Mitchell2019-hj}.

This position paper identifies ML research practices that differ from the methodologies of other scientific domains. We conjecture that grounding ML research in statistically sound hypothesis testing with careful control of nuisance parameters may encourage the publication of advances that stand the test of time. As in \citet{Rahimi2017-jz, Rahimi2018-se}, we consider methodological techniques in physics to suggest approaches that ML research scientists can adopt to ground their work in the conventions of empirical science. Proper application of the scientific method can help researchers understand factors of variation in experimental outcomes, as well as the dynamics of components in ML models, which would aid in ensuring robust performance in real world systems.

\textit{Explorimentation}, or the practice of poking around to see what happens, while appropriate for the early stages of research to inform and guide the formulation of a plausible hypothesis, does not constitute sufficient progress to term the effort scientific. 
As the field evolves towards a more mature state, a pragmatic approach (``It works!") is no longer sufficient. It is necessary to get past the stage of exploratory analysis and start venturing into rigorous hypothesis formulation and testing.

\section{The Scientific Method}
Starting from the assumption that there exists accessible ground truth, the scientific method is a systematic framework for experimentation that allows researchers to make objective statements about phenomena and gain knowledge of the fundamental workings of a system under investigation. 
The scientific method can also be seen as a social contract, a set of conventions that the community of researchers agrees to follow to everyone's benefit, rather than the one and only path to knowledge. The reasons for its existence are practical.
The goal of adopting a universally accepted set of guidelines is to be able to gain confirmable knowledge and insight into patterns of behavior that can be modeled systematically and can be reproduced in an experimental environment, under the assumption of there being a fundamental cause that drives the observed phenomena.

Central to this framework is the formulation of a scientific hypothesis and an expectation that can be falsified through experiments and statistical methods \citep{Jeffrey1956-uw, Popper1959-by, Lindberg1967-se, Arbuthnot1710-vi, Student1908-mb}; failure to include these steps is likely to lead to unscientific findings.
A well-formed hypotheses generates logical predictions and motivates experimentation: ``If the hypothesis is right, then I should expect to observe..." 
Rather than relying on untested speculations after the experiment has been conducted,
both the null and alternative hypotheses ought to be stated prior to collecting data and statistical testing.
Examples of hypothesis statements in the literature include those in \citet{Nguyen2015-pz, Azizzadenesheli2018-gq, Recht2019-km}.

At the base of scientific research lies the notion that an experimental outcome is a random variable, and that appropriate statistical machinery must be employed to estimate the properties of its distribution. 
It is important not to forget the role of chance in measurements: ``What are the odds of observing this result if the hypothesized trend is not real and the result is simply due to a statistical fluctuation?"
The first step towards a scientific formulation of ML then demands a more dramatic shift in priorities from drawing and recording single instances of experimental results to collecting enough data to gain an understanding of population statistics. 
Since abundant sampling of observations might be prohibitive due to resource constraints, the role of statistical uncertainties accompanying the measurement becomes vital to interpret the result. 

\citet{Neal1998-ik, Dietterich1998-ay,  Nadeau2000-hh, Bouckaert2004-bl} provided ML researchers with methods of statistical comparison across experimental results on supervised tasks. \citet{Demsar2006-cp} proposes methods for comparing two or more classifiers across multiple datasets, but also observes a lack of consensus in statistical testing methods among ICML papers.  These methods are not ubiquitous means of comparison between ML models and may not be applicable to all research questions. \citet{Henderson2018-yx} measure variation in performance in deep reinforcement learning (RL) models across parameter settings, and provides statistical tests to accompany some of these measurements, but they acknowledge that deep RL does not have common practices for statistical testing.

One may argue that the emphasis on hypotheses and statistical testing has created strong incentives in other fields to produce results that show $p\leq0.05$. These fields face a reproducibility crisis \citep{Ioannidis2005-yd, Baker2016-gp} and discourage negative results \citep{Rosenthal1979-zs, Brodeur2016-yp, Schuemie2018-zr}. To ensure that these hypotheses are created prior to the start of an experiment, researchers developed systems for pre-registration of hypotheses \citep{Dickersin2003-et, Nosek2018-pj}. One of the most notable systems is \texttt{clinicaltrials.gov}, which contains information on 300 thousand clinical trials occurring in 208 countries; registration is required for U.S. trials. In at least 165 journals, pre-registered studies may publish results, positive or negative, as a peer-reviewed registered report \citep{cos-rr, Nosek2014-hz, Chambers2014-sm}. A study of 127 registered reports in biomedicine and psychology found that 61\% published negative results. Researchers in ML have called for the publication of negative results that are grounded careful experimentation, and such publication systems may contribute to this shift in research practices.

\section{Case Study: Experimental High Energy Physics}
As reference, we offer insight into experimental high energy physics (HEP) --- one of many experimental fields in which the scientific method, based on hypothesis formulation and testing, is widely applied. While there has been a long history of statistical physics being utilized in machine learning, such as  \citet{Ackley1985-en} and \citet{Geman1984-it}, the methodological practices of experimental physics have also been suggested as a model for improved rigor in ML \citep{Rahimi2017-jz}. \citet{Platt1964-do} specifically identifies HEP as an area of science with notable adherence to hypotheses, a practice he causally links to the rapid advances made possible in that field.  Despite the chronological, technical, and methodological differences between ML and HEP, there exist general principles that transcend discipline boundaries and form the necessary components of any rational inquiry that wants to get elevated to the status of science.

In HEP, to check for discovery, the rationale followed to test the validity of a proposed theory (alternative hypothesis) is to compare its observable predictions to those made under a Standard Model-only~\citep{PhysRevD.98.030001} assumption (null hypothesis). Systematic uncertainties enter the estimation of observable effects under both hypotheses, and their careful accounting, along with that of statistical sources of uncertainty, can inform researchers of the expected sensitivity of their analysis before any data is collected. 
Typically, in HEP, the statistical procedure consists of two steps: the model building phase, and the hypothesis testing phase.

Analytical, parametric models can be derived from first principles or from fits to distributions of observables.
The measurement of parameters of interest $\mu$ is performed by maximizing the likelihood of the data under the model, while accounting for deviations that can be explained via nuisance parameters $\nu$.
Nuisance parameters are allowed to fluctuate within their uncertainties to represent the degree of uncertainty with which their systematic effect on the measurement is known, and provide slack to the fit.
They are incorporated in an extended likelihood via multiplicative factors: ${L}(\mu, \nu) = P_x(x; \mu, \nu) P_z(z; \nu)$, where the measurements of $\mu$ and $\nu$ occur in statistically orthogonal datasets $x\sim X$ and $z\sim Z$.
The portion of the likelihood associated with nuisance parameters estimates the trade-offs associated with moving away from their nominal values in order for the fit to converge. If a nuisance parameter is known with high accuracy, large deviations from its mean will make the solution less probable. 
High systematic uncertainties reduce the ability to narrow down the plausible values of $\mu$ --- many will correspond to similarly high values of the likelihood given the freedom to tune the nuisance portion of the model accordingly, which reduces the sensitivity of the experiment. This is the regime in which we expect many ML experiments to live.

A statistical test is run by constructing a suitable test statistic that is solely a function of the value of $\mu$ being tested.
Manually scanning over values of $\mu$, or relying on asymptotic properties of the test statistic, allows to construct frequentist Neyman confidence intervals~\citep{Neyman:1937uhy} and identify a range of $\mu$ values for which the $p$-value remains lower than a pre-determined magnitude for all values of the nuisance parameters. More details are offered in Appendix~\ref{appendix}.

\subsection{Strict Analogy for Machine Learning Experiments}
As physicists compute expectations over the number of events in the differential distribution of some discriminative observable, similarly, ML scientists could entertain the idea of investigating neural networks as physical objects and their time evolution as a natural phenomenon that follows the laws of dynamics, modeling ML experimental outcomes as variables, and analyzing them within a HEP-like hypothesis formulation and testing tradition. 

For instance, assume one postulated that a new activation function would intervene on information and gradient propagation in a specific, desirable way. First, they would formulate a quantitative, informed hypothesis of the expected behavior of a model with and without the suggested change. Then, if not already available from prior measurements, they would record outcomes from a variety of baseline models on a variety of reasonable datasets of similar characteristics in statistically orthogonal scenarios from the ones in which the measurement is being made, so as to constrain the nuisance parameters. Next, following a simplified additive model, they could assume, for example, that the observed performance using the proposed architectural modification may be separated as follows: (baseline performance) + $\mu \times$ (expected performance improvement from new activation), where both terms on the right hand side can be affected by nuisance parameters. Examples of nuisance parameters in ML experiments include: the choice of dataset, optimizer, initialization, hyperparameters, activation, normalization, regularization. Their estimation and modeling is not always an exact science and judgment calls will happen~\citep{sinervo2003definition}; before the field converges to well-calibrated, agreed-upon uncertainty modeling procedures, coarse and conservative decisions can suffice.  However, without appropriate effort from the community in handling and reducing systematic uncertainties from hyperparameter and experimental setup variations, it is unlikely for any analysis to have any sensitivity to the parameter of interest $\mu$. Indeed, since many current ML publications omit this fundamental step, it is plausible that a significant percentage of published work claiming state-of-the-art performance actually has no statistical sensitivity to measure their improvement over competing methods. Indeed, one would expect a large range of values of $\mu$, including 0, to be compatible with a high likelihood value, given the poor constraints on systematics. While reproducibility seeks to ensure \enquote{the ability of an experiment to be repeated with minor differences from the
original experiment, while achieving the
same qualitative results} \citep{Nagarajan2019-vd}, this goal on its own does not explicitly measure the systematic uncertainty of experimental results. 

In addition, borrowing another analogy from the field of physics, given the infancy of machine learning as a science, we would like to dissuade researchers from feeling the need to immediately have to come up with and test an all-encompassing ``Theory of Everything," and instead focus on measurements of first order behaviors first, as done in \citet{Shallue2018-ok}.  In loose analogy to perturbation theory in physics, one can study phenomena at different orders in an approximate expansion, starting from main trends and low order effects, then expanding to include higher order interaction terms. In other words, we first need to have a sense of how, say, BatchNorm~\citep{ioffe2015batch} roughly affects convergence, then model the interaction of it with depth, width, and skip connections~\citep{he2016deep}, among others.  

\section{Conclusions and Recommendations}
Hypothesis formulation prior to experimentation and statistical testing are two central pillars of the scientific method which are extremely rarely explicitly found in contemporary machine learning research papers. We strongly suggest that researchers incorporate these stages of experimentation into their work, perhaps by drawing inspiration, when possible, from the methodologies devised by other scientific disciplines. While rejecting mere scientism, we argue that it is necessary for the field to adopt the methodological tooling of empiricism and naturalism by operating in controlled, reproducible, and verifiable settings.
While these practices are primarily targeted to researchers interested in fundamental science of deep learning, applied researchers and research engineers will also benefit from the birth of a more principled, scientific subfield of machine learning.

Workshops, conferences, and panels should make the effort to include scientists, philosophers of science, and historians of science in conversations around the necessary steps to favor the transition of deep learning to a science. 

In addition, in the spirit of the Workshop on Negative Results in Computer Vision at CVPR 2017, we support the proposal of a future workshop which uses a registered reports model \citep{cos-rr}. Testable hypotheses would be submitted for approval ahead of time, and resulting contributions will be accepted regardless of their results with respect to accepting or rejecting the null hypothesis, granted that reviewers deem the authors' methods to be technically and scientifically sound. 

Finally, we invite the community of reviewers to pay closer attention to the accounting of statistical and systematic uncertainties which plague many state-of-the-art results, and consider the scientific robustness of claims. Submissions should not be discouraged for conflicting with other results, as long as prior art is acknowledged and confronted, and the authors are explicit about their result being incompatible with other findings. As \citet{gauch2003scientific} state, citing \citet{Megill1994-MEGRO-2},\enquote{objective truth expresses beliefs  `towards which all inquirers of good will are destined to converge.'}

\subsubsection*{Acknowledgments}
The authors thank Devi Parikh, L\'eon Bottou, Luke de Oliveira, Mustafa Mustafa, Ben Nachman, and Matthew Feickert for useful feedback.

\bibliography{citations}
\bibliographystyle{iclr2019_conference}

\appendix
\section{Additional Information on Hypothesis Testing in HEP}
\label{appendix}

A commonly adopted statistic in HEP is the profile likelihood-ratio $\lambda(\mu) = \frac{{L}(\mu, \hat{\hat{\nu}}(\mu))}{{L}(\hat{\mu}, \hat{\nu})} = \frac{\max_\nu {L}(\mu, \nu)}{{L}(\hat{\mu}, \hat{\nu})}$, in which nuisance parameters $\nu$ are profiled, \textit{i.e.} at the numerator, they are first expressed as functions of $\mu$ and maximized to remove any dependence on them ($\hat{\hat{\nu}}(\mu)$ is the conditional maximum likelihood estimator of a nuisance parameter), while the denominator represents the maximum of the unconstrained likelihood and is jointly maximized over $\mu$ and $\nu$ (the single-hat notation indicates the maximum likelihood estimator of a parameter).

The $p$-value
\begin{equation}
p_\mu = \int_{t_{\mu,\ \text{obs.}}}^{\infty} f(t_\mu|\mu)dt_\mu
\end{equation}
can be expressed in terms of the reparametrized profile likelihood ratio $t_\mu = -2 \ln \lambda(\mu)$, which enjoys better numerical stability.
In the asymptotic limit, $f(t_\mu)$ is $\chi^2$ distributed~\citep{wilks1938large}.

Incidentally, HEP adopts a unique approach to persistency of analyses: analysis preservation and combination are primarily achieved by sharing serialized versions of the likelihood model~\citep{verkerke2006roofit}, though solutions for analysis code sharing and data preservation have recently gained more traction in the community~\citep{Buckley:2010ar, Maguire:2017ypu, cowton2015open}.

We refer interested readers to the extensive prior literature for important considerations and information about advanced statistical methods and topics commonly used in HEP, such as the CLs method~\citep{Read:2002hq} (where the probability for the signal + background hypothesis is normalized by the background-only probability to avoid spurious exclusions of the null hypothesis in low experimental sensitivity analyses) or the ``Look-Elsewhere Effect"~\citep{Gross2010} (which points to the need for accounting for large parameter spaces and numbers of trials when assessing the true statistical significance of an observation).

\end{document}